\documentclass[sigconf, nonacm]{acmart}

\usepackage[inkscapelatex=false]{svg}
\usepackage{booktabs}
\usepackage{float}
\usepackage{multirow}
\usepackage{booktabs} 
\usepackage{array}    
\usepackage[table]{xcolor}
\usepackage{tabularx}
\usepackage{booktabs} 
\usepackage[most]{tcolorbox}
\usepackage{enumitem}
\usepackage{graphicx}
\usepackage[utf8]{inputenc}
\usepackage{pifont}
\usepackage{tikz}
\usetikzlibrary{matrix, calc}
\usepackage{balance}

\definecolor{stanfordred}{HTML}{B02418}

\definecolor{navyblue}{HTML}{000080}

\newcommand{\mycirc}[1]{%
  \tikz[baseline=-0.65ex]{%
    \node[shape=circle, fill=stanfordred, text=white, draw=black, thin, inner sep=0.5pt, minimum size=0.9em] (char) {\bfseries\footnotesize #1};%
  }%
}
\newcommand\vldbdoi{XX.XX/XXX.XX}
\newcommand\vldbpages{XXX-XXX}
\newcommand\vldbvolume{19}
\newcommand\vldbissue{12}
\newcommand\vldbyear{2026}
\newcommand\vldbauthors{\authors}
\newcommand\vldbtitle{\shorttitle} 
\newcommand\vldbavailabilityurl{https://github.com/asmath472/GraphContainer}
\newcommand\vldbpagestyle{plain} 

\begin{document}
\title{GraphContainer: A Unified Platform for Comparing and Debugging Graph RAG Methods}

\author{Seonho An}
\orcid{0009-0008-9280-0254}
\authornote{Co-first author.}
\affiliation{%
  \institution{KAIST}
  \country{Daejeon, Republic of Korea}
}
\email{asho1@kaist.ac.kr}

\author{Chaejeong Hyun}
\orcid{0009-0007-8349-4868}
\authornotemark[1]
\affiliation{%
  \institution{KAIST}
  \country{Daejeon, Republic of Korea}
}
\email{hchaejeong@kaist.ac.kr}

\author{Min-Soo Kim}
\authornote{Corresponding author.}
\orcid{0000-0002-5065-0226}
\affiliation{%
  \institution{KAIST}
  \country{Daejeon, Republic of Korea}
}
\email{minsoo.k@kaist.ac.kr}

\begin{abstract}
Graph RAG mitigates hallucinations and stale knowledge in LLMs, particularly for multi-hop question answering.
However, existing approaches remain highly fragmented and incompatible. The structural heterogeneity of graph formats across different frameworks and the lack of granular visualization tools make it exceedingly difficult to evaluate and compare retrieval behaviors.
To bridge this gap, we propose \textbf{GraphContainer}, a novel platform designed to unify and visualize diverse graph RAG workflows.
GraphContainer features two key components: (1) a \textbf{Unified Graph Representation (UGR) layer} that seamlessly standardizes multi-format graphs, and (2) a \textbf{Graph Recorder} that tracks and visually renders the step-by-step retrieval process.
Through an interactive web interface, we demonstrate GraphContainer's ability to import heterogeneous graphs and perform live, traceable visual debugging of graph RAG methods.
Ultimately, we show how GraphContainer enables controlled comparisons of various graph formats and retrieval strategies, lowering the barrier for researchers and practitioners to design optimal graph RAG pipelines. 
A demonstration video is available at \href{https://youtu.be/O02eNJLwkU0}{https://youtu.be/O02eNJLwkU0}.
\end{abstract}

\maketitle

\pagestyle{\vldbpagestyle}
\begingroup\small\noindent\raggedright\textbf{PVLDB Reference Format:}\\
\vldbauthors. \vldbtitle. PVLDB, \vldbvolume(\vldbissue): \vldbpages, \vldbyear.\\
\href{https://doi.org/\vldbdoi}{doi:\vldbdoi}
\endgroup
\begingroup
\renewcommand\thefootnote{}\footnote{\noindent
This work is licensed under the Creative Commons BY-NC-ND 4.0 International License. Visit \url{https://creativecommons.org/licenses/by-nc-nd/4.0/} to view a copy of this license. For any use beyond those covered by this license, obtain permission by emailing \href{mailto:info@vldb.org}{info@vldb.org}. Copyright is held by the owner/author(s). Publication rights licensed to the VLDB Endowment. \\
\raggedright Proceedings of the VLDB Endowment, Vol. \vldbvolume, No. \vldbissue\ %
ISSN 2150-8097. \\
\href{https://doi.org/\vldbdoi}{doi:\vldbdoi} \\
}\addtocounter{footnote}{-1}\endgroup

\ifdefempty{\vldbavailabilityurl}{}{
\vspace{.3cm}
\begingroup\small\noindent\raggedright\textbf{PVLDB Artifact Availability:}\\
The source code, data, and/or other artifacts have been made available at \url{\vldbavailabilityurl}.
\endgroup
}
\section{Introduction}
\label{sec:intro}

Retrieval-augmented generation (RAG) methods have emerged as a foundational technology for addressing the hallucination and out-of-date information problems inherent in large language models (LLMs)~\cite{gao2023retrieval}.
Recently, graph RAG methods—which leverage structured graph data into the retrieval pipeline—demonstrate superior effectiveness, particularly in multi-hop question answering~\cite{InDepthGraphRAG-Zhou-VLDB-2025}.

The proliferation of graph RAG methods poses a challenge in identifying the best approach for specific datasets or tasks~\cite{lego-graphrag}, a research area that remains largely unexplored due to two fundamental bottlenecks.
First, comparative analysis is severely hindered by \textbf{structural heterogeneity}. 
As shown in Table \ref{tab:format_notations}, representative graph formats utilize highly divergent data notation. Since these formats are tightly coupled to their native graph RAG method, it is difficult to cross-evaluate methods across datasets. 
Second, \textbf{analyzing the graph retrieval process is elusive}. 
Tracking the step-by-step selection of nodes and edges during a query requires dedicated visualization tools. 
Although some methods (e.g. LightRAG~\cite{lightrag}) offer built-in visualizers, they are strictly isolated to their own architectures. 
Consequently, researchers and practitioners lack a cohesive environment to compare, evaluate, and debug diverse graph RAG strategies~\cite{lego-graphrag}.

\begin{figure*}[tb!]
\centering
\includegraphics[width=\textwidth]{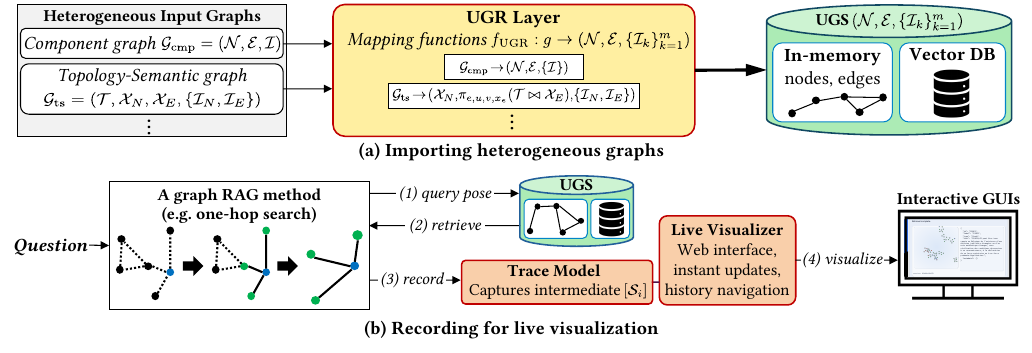}
\vspace{-0.9cm}
\caption{Overview of GraphContainer. (a) UGS transformation via UGR. (b) Live visualization for Graph RAG.}
\vspace{-0.3cm}
\label{fig:overview}
\end{figure*}

\begin{table}[htbp]
\centering
\caption{Notation of four representative graph formats.
We use $u,v$ for node identifiers, $x_u$ for node content, $\mathbf{z}_u \in \mathbb{R}^d$ for node embeddings,
$e$ for edge identifiers, $x_e$ for edge content, and $\mathbf{z}_e \in \mathbb{R}^d$ for edge embeddings.}
\vspace{-0.2cm}
\label{tab:format_notations}
\resizebox{\columnwidth}{!}
{
\begin{tabular}{ll}
\toprule
\textbf{Graph Format} & \textbf{Notation} \\
\midrule
Component Graph~\cite{fastinsight} &
\begin{tabular}[c]{@{}l@{}}
$\mathcal{G}_{\mathrm{cmp}} = (\mathcal{N}, \mathcal{E}, \mathcal{I})$ where $\mathcal{N}=\{(u, x_u)\}$, \\
$\mathcal{E}=\{(u, v)\}$, $\mathcal{I}=\{(u, \mathbf{z}_u)\}$
\end{tabular} \\
\midrule
Attribute Bundle Graph~\cite{lightrag, pathrag} &
\begin{tabular}[c]{@{}l@{}}
$\mathcal{G}_{\mathrm{bun}} = (\mathcal{N}^{\ast}, \mathcal{E}^{\ast})$ where \\ $\mathcal{N}^{\ast}=\{(u, x_u, \mathbf{z}_u)\}$, $\mathcal{E}^{\ast}=\{(e, u, v, x_e, \mathbf{z}_e)\}$
\end{tabular} \\
\midrule
Topology--Semantic Graph~\cite{hipporag2} &
\begin{tabular}[c]{@{}l@{}}
$\mathcal{G}_{\mathrm{ts}} = (\mathcal{T}, \mathcal{X}_N, \mathcal{X}_E, \{\mathcal{I}_N, \mathcal{I}_E\})$, where \\ 
$\mathcal{T}=\{(e, u, v)\}$, $\mathcal{X}_N=\{(u, x_u)\}$, $\mathcal{X}_E=\{(e, x_e)\}$ \\
$\mathcal{I}_N=\{(u, \mathbf{z}_u)\}$, $\mathcal{I}_E=\{(e, \mathbf{z}_e)\}$
\end{tabular} \\
\midrule
Subgraph Union Graph~\cite{g-retriever, grag, kg2rag} &
\begin{tabular}[c]{@{}l@{}}
$\mathcal{G}_{\mathrm{sub}} = \{(\mathcal{N}_i, \mathcal{E}_i, \{\mathcal{I}_{N,i}, \mathcal{I}_{E,i}\})\}_{i=1}^{m}$ \\
where $\mathcal{N}_i=\{(u, x_u)\}$, $\mathcal{E}_i=\{(e, u, v)\}$ \\
$\mathcal{I}_{N,i}=\{(u, \mathbf{z}_u) | u \in \mathcal{N}_i\}$, $\mathcal{I}_{E,i}=\{(e, \mathbf{z}_e)|e \in \mathcal{E}_i\}$
\end{tabular} \\
\bottomrule
\end{tabular}
}
\end{table}


To bridge this gap, we propose \textbf{GraphContainer}, a novel platform for the comparison and visual debugging of graph RAG methods. 
GraphContainer addresses the above challenges through two key components: 
(1) a \textbf{Unified Graph Representation (UGR) layer} and (2) a \textbf{Graph Recorder}.
First, the UGR layer maps disparate graph formats (e.g., four major formats detailed in Table~\ref{tab:format_notations}) into a standardized structure called the \textbf{Unified Graph State (UGS)}. The UGS comprehensively captures both structural and semantic attributes across all formats, enabling the underlying retrievers to operate over a shared representation. 
Second, the \textbf{Graph Recorder} tracks the step-by-step execution of retrieval methods operating on the UGS. 
This recorded process is then rendered through our Live Visualizer, enabling users to inspect exact retrieval behaviors and interactively refine their graph RAG pipelines.

We demonstrate GraphContainer's capabilities through an end-to-end scenario where users import heterogeneous graphs into a unified representation and visualize the entire retrieval process via a live web interface.
Finally, we present an empirical study illustrating how GraphContainer facilitates controlled, rigorous comparisons across varying graph construction and retrieval methods.


\section{System Overview}
\label{sec:system}

As illustrated in Figure~\ref{fig:overview}, our GraphContainer provides two core functionalities: (a) the ingestion and standardization of heterogeneous graphs via the UGR layer, and (b) dynamic graph tracking and live visualization via Graph Recorder.
Figure~\ref{fig:overview}(a) depicts functionality (a), where the UGR layer employs format-specific mapping functions to decode varying graph structures into a single canonical UGS. 
Meanwhile, for functionality (b), the Graph Recorder captures the dynamic execution state of the retrievers operating over this shared UGS. Figure~\ref{fig:overview}(b) shows the step-by-step process (1--4) that translates these recorded results into a live visualization within the graph RAG workflow.


\subsection{Unified Graph Representation Layer}
\label{subsec:ugr}

Existing graph RAG methods encode structural knowledge using diverse \textit{graph formats}.
Formally, a graph format is defined as:
\begin{definition}[Graph Format]
    A graph format $\mathcal{G}$ is defined as a data schema that encodes nodes $\mathcal{N}$, edges $\mathcal{E}$, and $m$ optional vector indices $\{\mathcal{I}_k\}_{k=1}^m$.
\end{definition}
These encoded elements serve as fundamental data components for retrieval in graph RAG methods~\cite{InDepthGraphRAG-Zhou-VLDB-2025, fastinsight}.

Although various graph formats hold logically equivalent information, their structural discrepancies make systematic comparison of graph RAG methods difficult.
To resolve this, we propose the \textbf{Unified Graph Representation (UGR)} layer. 
The UGR layer acts as a translator, taking disparate graph formats and standardizing them into a common structure referred to as the \textbf{Unified Graph State (UGS)}. 
Here, the UGS serves as the materialized representation of $\mathcal{N}$, $\mathcal{E}$, and $\{\mathcal{I}_k\}_{k=1}^m$.
Nodes, edges, and their attributes are maintained in the \textit{in-memory UGS}, while vector indices are managed via an external vector database.
Formally, the UGR layer is defined as follows:
\begin{definition}[Unified Graph Representation Layer]
\begin{sloppypar}
    Given a graph instance $g$ encoded in format $\mathcal{G}$, the UGR is defined by a set of mapping functions, $f_{\text{UGR}} : g \rightarrow (\mathcal{N}, \mathcal{E}, \{\mathcal{I}_k\}_{k=1}^m)$, that transform format-specific components into the UGS.
\end{sloppypar}
\end{definition}

In practice, our proposed UGR layer supports four major graph formats categorized from representative graph RAG frameworks:
(1) \textit{Component Graph} in FastInsight~\cite{fastinsight},
(2) \textit{Attribute Bundle Graph} in LightRAG~\cite{lightrag} and PathRAG~\cite{pathrag},
(3) \textit{Topology-Semantic Graph} in HippoRAG~\cite{hipporag2},
(4) \textit{Subgraph Union Graph} in G-Retriever~\cite{g-retriever}, GRAG~\cite{grag}, and KG2RAG~\cite{kg2rag}.
Although we initially focus on these four formats, users can easily extend this to custom graph formats by adding corresponding UGR mapping functions.
We summarize the format-specific decoding mappings induced by $f_{\text{UGR}}$ in Table~\ref{tab:decoding_logic}.

\begin{table}[htbp]
\centering
\vspace{-0.2cm}
\caption{Summary of decoding mappings from each graph format to the UGS. Here, $\pi$ denotes projection, $\bowtie$ denotes join, and $\phi_N, \phi_E$ denote parsing functions that convert textual triplets into node and edge records.}
\vspace{-0.3cm}
\label{tab:decoding_logic}
\resizebox{\columnwidth}{!}
{
\begin{tabular}{ll}
\toprule
\textbf{Graph Format} & \textbf{Decoding Mapping} \\
\midrule
Component Graph & $(\mathcal{N}, \mathcal{E}, \{\mathcal{I}\})$ \\
\midrule
Attribute Bundle Graph & $\left(\pi_{u, x_u}(\mathcal{N}^{\ast}),\!\pi_{e, u, v, x_e}(\mathcal{E}^{\ast}),\!\{\pi_{u, \mathbf{z}_u}(\mathcal{N}^{\ast}),\!\pi_{e, \mathbf{z}_e}(\mathcal{E}^{\ast})\}\right)$ \\
\midrule
Topology--Semantic Graph & $\left(\mathcal{X}_N,\; \pi_{e,u,v,x_e}(\mathcal{T} \bowtie \mathcal{X}_E),\; \{\mathcal{I}_N, \mathcal{I}_E\}\right)$ \\
\midrule
Subgraph Union Graph & $\left(\bigcup_{i=1}^{m} \mathcal{N}_i, \, \bigcup_{i=1}^{m} \mathcal{E}_i, \, \{ \bigcup_{i=1}^{m} \mathcal{I}_{N,i}, \bigcup_{i=1}^{m} \mathcal{I}_{E,i} \} \right)$ \\
\bottomrule
\end{tabular}
}
\vspace{-0.4cm}
\end{table}

\begin{figure*}[t!]
    \centering
    \includegraphics[width=\textwidth]{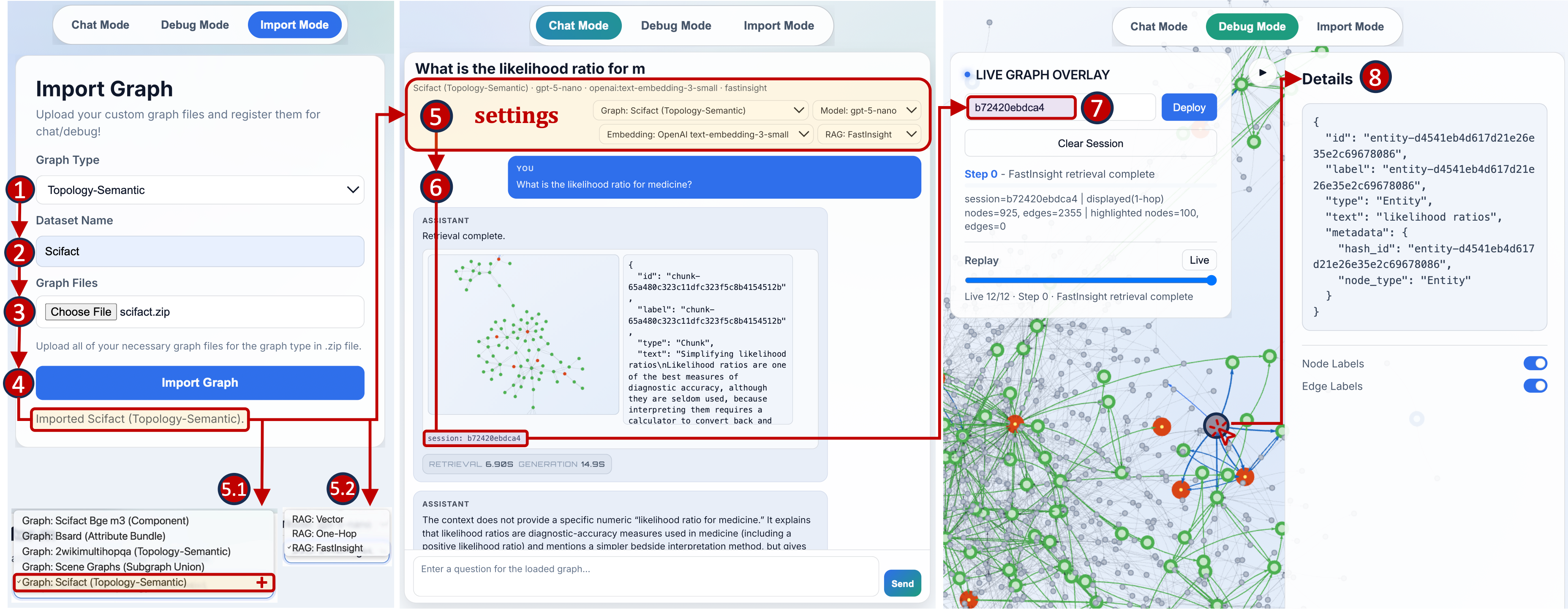}
    \vspace{-0.6cm}
    \caption{A demonstration flow for our GraphContainer.}
    \vspace{-0.4cm}
    \label{fig:demo}
\end{figure*}

\subsection{Graph Recorder}
\label{subsec:recorder}

The Graph Recorder is comprised of two components: a \textbf{Trace Model} that captures intermediate retrieval results (i.e., retrieved nodes and edges), and a \textbf{Live Visualizer} that renders them to analyze retrieval processes on the UGS. 

During retrieval, a query is first processed by a selected graph RAG method, which retrieves relevant node and edge identifiers from the UGS (steps (1)--(2) in Figure~\ref{fig:overview}(b)). The retriever then calls the Trace Model to organize each retrieval progress per query as a \textit{session} and record the current retrieval results (step (3)) by capturing the nodes and edges selected at each step. As retrieval progresses, these recorded results accumulate into a sequence $[\mathcal{S}_1, \mathcal{S}_2, \cdots, \mathcal{S}_m]$ for each session, where each $\mathcal{S}_i$ denotes the subgraph at the $i$-th record call induced by the retrieved nodes and edges in the in-memory UGS.
By recording each step as a subgraph $\mathcal{S}_i$ that stores only incremental overlay information, the Trace Model minimizes storage overhead while preserving the structural context required for inspection.

Next, the Live Visualizer consumes the recorded sequence produced by the Trace Model and renders it through a web-based interactive interface (step (4)). 
Each invocation of step (3) appends a new subgraph $\mathcal{S}_i$ to the sequence and triggers an immediate update of the visualized graph to reflect the latest retrieval step. The visualizer preserves this sequence as the retrieval history within a session, allowing users to navigate $\mathcal{S}_i$ and inspect how the retrieved subgraph evolves. For example, users can monitor the evolving retrieval process while interacting with standard debugging environments (e.g., Python debuggers), or modify the graph structure or retrieval logic to observe the changes immediately in the interface. 

\vspace{-0.1cm}
\section{Demonstration Scenarios}
\label{sec:demo}



We demonstrate the debuggability and comparability of GraphContainer through two scenarios: \textbf{Interactive Visual Debugging} and \textbf{Comparative Retrieval Analysis}. 
In the first scenario, users experience the end-to-end GraphContainer workflow by \textit{importing} a sample graph through the GUI, \textit{executing} Graph RAG queries, and \textit{visualizing} the step-by-step retrieval process. 
In the second scenario, users compare various retrieval methods---specifically vector search, one-hop search, and FastInsight---to systematically analyze how different retrieval mechanisms influence generated answers.

\vspace{-0.1cm}
\subsection{Scenario 1: Interactive Visual Debugging}
The first scenario consists of three sequential phases: Importing, Executing, and Visualizing.
These phases correspond to steps \mycirc{1}--\mycirc{4}, \mycirc{5}--\mycirc{6}, and \mycirc{7}--\mycirc{8} in Figure~\ref{fig:demo}, respectively.

\vspace{1mm}
\noindent \textbf{Importing.} In this phase, we provide users with two sample graphs: (a) BSARD in the Attribute Bundle format, and (b) SciFact in the Topology-Semantic format. 
To begin the demonstration session, users first navigate to the \textit{Import Mode} on the website. 
Users select one of the provided graphs and upload it to the GraphContainer platform.
Specifically, users select the format of the chosen graph (\mycirc{1} in Figure~\ref{fig:demo}), input the name of the dataset to be displayed (\mycirc{2}), upload the graph files (\mycirc{3}), and click the \textit{Import Graph} button (\mycirc{4}). Subsequently, the server utilizes UGR to decode the uploaded graph into the canonical UGS.

\vspace{0.5mm}
\noindent \textbf{Executing.} 
Switching to the \textit{Chat Mode} interface at the top of the web, users configure the graph RAG pipeline (\mycirc{5}) by selecting the target graph dataset (\mycirc{5.1}), the embedding and LLM models, and the retrieval method (\mycirc{5.2}). 
Users then submit a natural language query and review the generated response (\mycirc{6}). 
To facilitate the demonstration, we provide default queries (e.g., ``What do I risk if I violate professional confidentiality?'' for BSARD, and ``What is the likelihood ratio for medicine?'' for SciFact) and set FastInsight and one-hop search as the default retrieval methods. 
Users can also interactively submit their own custom queries.
Once the query completes, users can thoroughly inspect the underlying retrieval mechanics. The interface details the selected nodes (shown as red dots in the displayed subgraph), the retrieval and generation latencies, the final generated response, and the \textit{session\_id}.

\vspace{0.5mm}
\noindent \textbf{Visualizing.} In this final phase, users transition to the \textit{Debug Mode}. 
Upon submitting the \textit{session\_id} from the \textit{Executing} phase into the \textsc{Live Graph Overlay} panel and clicking \textit{Deploy}, the system dynamically renders the retrieved graph (\mycirc{7}).
Users can then interactively inspect specific node and edge attributes or utilize the replay slider to visualize the step-by-step retrieval sequence (\mycirc{8}). 

\vspace{-0.4cm}
\subsection{Scenario 2: Comparative Retrieval Analysis}
\label{subsec:compare-scenario}

In the second scenario, users execute and evaluate various RAG methods using three pre-defined retrievers: (1) vector search, which retrieves the top-$K$ nodes using an embedding model; 
(2) one-hop search, which expands the initial top-$K$ selection to include immediate neighboring nodes; and (3) FastInsight~\cite{fastinsight}. 

Returning to the\textit{ Chat Mode} interface, users execute an identical query while varying the selected retrieval method (\mycirc{5.2}).
By using the unique \textit{session\_id} from each execution, users can render and systematically contrast the resulting subgraphs in the \textit{Debug Mode}.
This process illustrates how GraphContainer seamlessly facilitates side-by-side comparative analysis across diverse graph RAG methods.

\section{Experimental Study}

Beyond the initial demonstration, we conduct an empirical study to illustrate the experiment types and analyses enabled by GraphContainer. 
Specifically, we investigate how different combinations of graph construction and retrieval choices affect overall graph RAG performance. 
While the heterogeneous formats produced by multiple graph RAG methods typically hinder direct comparisons, GraphContainer overcomes this challenge by unifying these representations into a common UGS.


\noindent \textbf{Settings.}
We utilize BSARD~\cite{bsard} and SciFact document corpora $(d_1, \dots, d_n)$ to construct experimental graphs. Specifically, we use three different construction methods: 
(1) LightRAG, which produces an \textit{Attribute Bundle Graph};
(2) HippoRAG, which produces a \textit{Topology-Semantic Graph}; and
(3) a standard k-NN approach that constructs a \textit{Component Graph} $(\mathcal{N}, \mathcal{E}, \{\mathcal{I}\})$ using document embeddings.
The UGR layer converts all graphs into UGS, enabling evaluation through a unified interface.


We evaluate two graph retrieval methods introduced in Section~\ref{subsec:compare-scenario}: \textit{One-hop search} and \textit{FastInsight}.
These retrievers are used in two corresponding RAG pipelines to generate answers from the retrieved information. 
The outputs are evaluated using an LLM-as-a-Judge protocol.
Overall, we evaluate six combinations (three graph construction methods $\times$ two graph RAG methods), resulting in 15 pairwise comparisons for each dataset.

For all experiments, we use OpenAI's text-embedding-3-small for embeddings, Gemma3 (12B) via Ollama for generation, OpenAI's GPT-5-mini as the judge model, and ChromaDB as the vector database.
To mitigate positional bias, each comparison is performed twice with reversed answer ordering, yielding a mean win ratio.
We use an initial retrieval size of ten and select the final five nodes in FastInsight; all other parameters follow the default settings.

\begin{table}[bp]
\centering
\vspace{-0.3cm}
\caption{Pairwise overall win rate. Top: BSARD, Bottom: SciFact. (KNN)/(HR)/(LR) denote $k$-NN/HippoRAG/LightRAG construction; FI/OH denote FastInsight/One-hop retrieval. Yellow cells indicate FI vs. OH for the same construction.}
\vspace{-0.4cm}
\label{tab:pairwise_win_rate}
\renewcommand{\arraystretch}{0.85}
\resizebox{\columnwidth}{!}
{
\begin{tabular}{l ccccc}
\toprule
& \multicolumn{5}{c}{\textbf{Win Rates (\%) of Column Method}} \\
\cmidrule(r){2-6}
\textbf{vs. Row Method} & (KNN)+FI & (KNN)+OH & (HR)+FI & (HR)+OH & (LR)+FI \\
\midrule
\multicolumn{6}{l}{\textbf{BSARD Dataset}} \\
(KNN) + OH & \cellcolor{yellow!50}59.01 & --    & --    & --    & --    \\
(HR) + FI  & 68.47 & 61.71 & --    & --    & --    \\
(HR) + OH  & 75.45 & 71.33 & \cellcolor{yellow!50}62.39 & --    & --    \\
(LR) + FI  & 95.72 & 93.92 & 87.39 & 83.56 & --    \\
(LR) + OH  & 95.05 & 88.74 & 84.91 & 81.31 & \cellcolor{yellow!50}50.23 \\
\midrule
\multicolumn{6}{l}{\textbf{SciFact Dataset}} \\
(KNN) + OH & \cellcolor{yellow!50}54.50 & --    & --    & --    & --    \\
(HR) + FI  & 63.06 & 57.21 & --    & --    & --    \\
(HR) + OH  & 71.17 & 66.22 & \cellcolor{yellow!50}57.66 & --    & --    \\
(LR) + FI  & 95.95 & 94.14 & 84.68 & 83.33 & --    \\
(LR) + OH  & 94.14 & 87.84 & 81.98 & 80.18 & \cellcolor{yellow!50}55.86 \\
\bottomrule
\end{tabular}
}
\end{table}

\noindent \textbf{Experimental Results and Discussions.}
Table~\ref{tab:pairwise_win_rate} reports the 30 pairwise comparisons and reveals two main findings. 
First, under the same graph construction (yellow cells in Table~\ref{tab:pairwise_win_rate}), FastInsight outperforms one-hop, though the margins of improvement exhibit considerable variance. 
Second, across both retrieval methods, the KNN-constructed graph yields the strongest overall performance on both datasets, followed by the HR graph.

These results suggest that the effectiveness of a graph RAG pipeline depends jointly on the interplay between the graph construction strategy and the retrieval method. More specifically, the significant performance variance between KNN and LLM-mediated restructuring methods (HR and LR) highlights the need for cross-evaluation. GraphContainer facilitates this process by enabling effortless comparisons, allowing researchers to thoroughly test robustness and practitioners to evaluate diverse pipelines with minimal engineering overhead.




\section{Conclusions and Future Works}

In this paper, we presented \textbf{GraphContainer}, a unified visual analytics platform designed to evaluate and debug graph RAG methods across heterogeneous formats. 
By integrating a \textbf{Unified Graph Representation} layer with an interactive \textbf{Graph Recorder}, GraphContainer facilitates standardized graph conversion alongside step-by-step inspection of retrieval behaviors.
Through interactive scenarios and an empirical study, we demonstrated the platform's capability to support controlled comparisons across different graph construction and retrieval methods. 
We expect GraphContainer to lower the barrier to adopting graph RAG by making complex workflows more transparent, comparable, and highly accessible.

For future work, we plan to transition from in-memory graph to a disk-based graph for massive, enterprise-scale support, and integrate LLM-driven automated graph RAG debugging to provide users with natural language explanations.




\bibliographystyle{ACM-Reference-Format}
\bibliography{sample}

\end{document}